\documentclass[11pt,a4paper]{article}
\usepackage[hyperref]{acl2020}
\usepackage{times}
\usepackage{latexsym}

\usepackage{amssymb}
\usepackage{amsmath}
\usepackage{graphicx}
\usepackage{caption}
\usepackage{subcaption}
\usepackage{comment}
\usepackage{multirow}
\usepackage{url}
\usepackage{tikz}
\usepackage{pgfplots}
\pgfplotsset{compat=1.14}

\usepackage{microtype}

\aclfinalcopy 

\title{End-to-End Neural Word Alignment Outperforms GIZA++}

\author{Thomas Zenkel, Joern Wuebker, John DeNero \\
   Lilt, Inc. \\
  {\tt first\_name@lilt.com}
}

\date{}

\begin{document}
\maketitle
\begin{abstract}

Word alignment was once a core unsupervised learning task in natural language processing because of its essential role in training statistical machine translation (MT) models.
Although unnecessary for training neural MT models, word alignment still plays an important role in interactive applications of neural machine translation, such as annotation transfer and lexicon injection. 
While statistical MT methods have been replaced by neural approaches with superior performance, the twenty-year-old GIZA++ toolkit remains a key component of state-of-the-art word alignment systems. 
Prior work on neural word alignment has only been able to outperform GIZA++ by using its output during training.
We present the first end-to-end neural word alignment method that consistently outperforms GIZA++ on three data sets. 
Our approach repurposes a Transformer model trained for supervised translation to also serve as an unsupervised word alignment model in a manner that is tightly integrated and does not affect translation quality.

\end{abstract}

\section{Introduction}
\label{sec:Introduction}
Although word alignments are no longer necessary to train machine translation (MT) systems, they still play an important role in applications of neural MT.
For example, they enable injection of an external lexicon into the inference process to enforce the use of domain-specific terminology or improve the translations of low-frequency content words \cite{Arthur_2016}.
The most important application today for word alignments is to transfer text annotations from source to target \citep{muller2017treatment, tezcan2011smt, joanistransferring, escartin2015machine}. For example, if part of a source sentence is underlined, the corresponding part of its translation should be underlined as well. HTML tags and other markup must be transferred for published documents. Although annotations could in principle be generated directly as part of the output sequence, they are instead typically transferred via word alignments because example annotations typically do not exist in MT training data.

The Transformer architecture provides state-of-the-art performance for neural machine translation \citep{vaswani2017attention}. The decoder has multiple layers, each with several attention heads, which makes it difficult to interpret attention activations as word alignments. As a result, the most widely used tools to infer word alignments, namely GIZA++ \citep{Och_2003} and FastAlign \citep{dyer2013simple}, are still based on the statistical IBM word alignment models developed nearly thirty years ago \citep{brown1993mathematics}. No previous unsupervised neural approach has matched their performance. Recent work on alignment components that are integrated into neural translation models either underperform the IBM models or must use the output of IBM models during training to outperform them \citep{zenkel2019adding,garg-etal-2019-jointly}.

This work combines key components from \citet{zenkel2019adding} and \citet{garg-etal-2019-jointly} and presents two novel extensions. Statistical alignment methods contain an explicit bias towards contiguous word alignments in which adjacent source words are aligned to adjacent target words. This bias is expressed in statistical systems using a hidden Markov model (HMM) \citep{vogel-etal-1996-hmm}, as well as symmetrization heuristics such as the grow-diag-final algorithm \citep{Och_2000,koehn2005edinburgh}.
We design an auxiliary loss function that can be added to any attention-based network to encourage contiguous attention matrices. 

\begin{figure}
    \centering
    \includegraphics[scale=0.28]{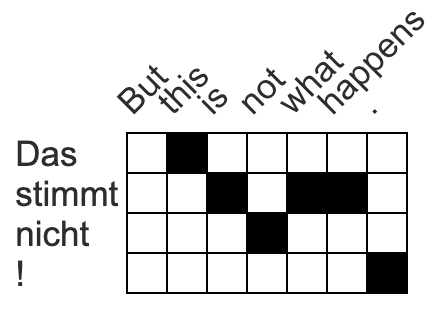}
    \caption{Word alignment generated by a human annotator.} 
    \label{fig:exampleGold}
\end{figure}

The second extension replaces heuristic symmetrization of word alignments with an activation optimization technique. 
After training two alignment models that translate in opposite directions, we infer a symmetrized attention matrix that jointly optimizes the likelihood of the correct output words under both models in both languages. 
Ablation experiments highlight the effectiveness of this novel extension, which is reminiscent of agreement-based methods for statistical models \citep{liang2006alignment,graca2008,denero2011model}.

End-to-end experiments show that our system is the first to consistently yield higher alignment quality than GIZA++ using a fully unsupervised neural model that does not use the output of a statistical alignment model in any way. 

\section{Related Work}

\label{sec:RelatedWork}
\subsection{Statistical Models}
Statistical alignment models directly build on the lexical translation models of \citet{brown1993mathematics}, known as the IBM models. The most popular statistical alignment tool is GIZA++ \citep{Och_2000,Och_2003, gao2008parallel}. For optimal performance, the training pipeline of GIZA++ relies on multiple iterations of IBM Model 1, Model 3, Model 4 and the HMM alignment model \citep{vogel-etal-1996-hmm}. Initialized with parameters from previous models, each subsequent model adds more assumptions about word alignments. Model 2 introduces non-uniform distortion, and Model 3 introduces fertility. Model 4 and the HMM alignment model introduce relative distortion, where the likelihood of the position of each alignment link is conditioned on the position of the previous alignment link.
While simpler and faster tools exist such as FastAlign \citep{dyer2013simple}, which is based on a reparametrization of IBM Model 2, the GIZA++ implementation of Model 4 is still used today in applications where alignment quality is important. 

In contrast to GIZA++, our neural approach is easy to integrate on top of an attention-based translation network, has a training pipeline with fewer steps, and leads to superior alignment quality. Moreover, our fully neural approach that shares most parameters with a neural translation model can potentially take advantage of improvements to the underlying translation model, for example from domain adaptation via fine-tuning.

\subsection{Neural Models}
Most neural alignment approaches in the literature, such as \citet{Tamura_2014} and \citet{alkhouli2018alignment}, rely on alignments generated by statistical systems that are used as supervision for training the neural systems. These approaches tend to learn to copy the alignment errors from the supervising statistical models.

\citet{zenkel2019adding} use attention to extract alignments from a dedicated alignment layer of a neural model without using any output from a statistical aligner, but fail to match the quality of GIZA++.

\citet{garg-etal-2019-jointly} represents the current state of the art in word alignment, outperforming GIZA++ by training a single model that is able to both translate and align. This model is supervised with a guided alignment loss, and existing word alignments must be provided to the model during training. \citet{garg-etal-2019-jointly} can produce alignments using an end-to-end neural training pipeline guided by attention activations, but this approach underperforms GIZA++. The performance of GIZA++ is only surpassed by training the guided alignment loss using GIZA++ output. Our method also uses guided alignment training, but our work is the first to surpass the alignment quality of GIZA++ without relying on GIZA++ output for supervision.

\citet{stengel-eskin-etal-2019-discriminative} introduce a discriminative neural alignment model that uses a dot-product-based distance measure between learned source and target representation to predict if a given source-target pair should be aligned. Alignment decisions condition on the neighboring decisions using convolution. The model is trained using gold alignments.  In contrast, our approach is fully unsupervised; it does not require gold alignments generated by human annotators during training.  
Instead, our system implicitly learns reasonable alignments by predicting future target words as part of the translation task, but selects attention activations using an auxiliary loss function to find contiguous alignment links that explain the data.

\section{Background}
\label{sec:Background}
\subsection{The Alignment Task}
Given a source-language sentence  $x = x_1, \dots, x_n$ of length $n$ and its target-language translation $y = y_1, \dots, y_m$ of length $m$, an alignment $\mathcal{A}$ is a set of pairs of source and target positions:
\begin{align*}
    \mathcal{A} \subseteq \{ (s, t): s \in \{ 1, \dots, n \}, t \in \{ 1, \dots, m \} \}
\end{align*}
Aligned words are assumed to correspond to each other, i.e. the source and the target word are translations of each other within the context of the sentence. Gold alignments are commonly generated by multiple annotators based on the Blinker guidelines \citep{melamed1998annotation}. 
The most commonly used metric to compare automatically generated alignments to gold alignments is alignment error rate (AER) \citep{Och_2000}.

\subsection{Attention-Based Translation Models}
\label{subsec:translationModel}
\citet{bahdanau2014neural} introduced attention-based neural networks for machine translation. These models typically consist of an encoder for the source sentence and a decoder that has access to the previously generated target tokens and generates the target sequence from left to right. Before predicting a token, the decoder ``attends'' to the position-wise source representations generated by the encoder, and it produces a context vector that is a weighted sum of the contextualized source embeddings.

The Transformer \citep{vaswani2017attention} attention mechanism uses a query $Q$ and a set of $k$ key-value pairs $K, V$ with $Q \in \mathbb{R}^{d}$ and $V, K \in \mathbb{R}^{k \times d}$. Attention logits $A_L$ computed by a scaled dot product are converted into a probability distribution $A$ using the softmax function. The attention $A$ serves as mixture weights for the values $V$ to form a context vector $c$:
\begin{align*}
    A_L &= \mathsf{calcAttLogits}(Q, K) = \frac{Q \cdot K^{T}}{\sqrt{d}} \\
    A &= \mathsf{calcAtt}(Q, K) = \mathsf{softmax}(A_L) \\
    c &= \mathsf{applyAtt}(A, V) = A \cdot V
\end{align*}

A state-of-the-art Transformer includes multiple attention heads whose context vectors are stacked to form the context activation for a layer, and the encoder and decoder have multiple layers. For all experiments, we use a downscaled Transformer model trained for translation with a 6-layer encoder, a 3-layer decoder, and 256-dimensional hidden states and embedding vectors.

For the purpose of word alignment, this translation Transformer is used as-is to extract representations of the source and the target sequences, and our alignment technique does not change the parameters of the Transformer.  Therefore, improvements to the translation system can be expected to directly carry over to alignment quality, and the alignment component does not affect translation output in any way.

\subsection{Alignment Layer}
\begin{figure}
    \centering
    \includegraphics[scale=0.33]{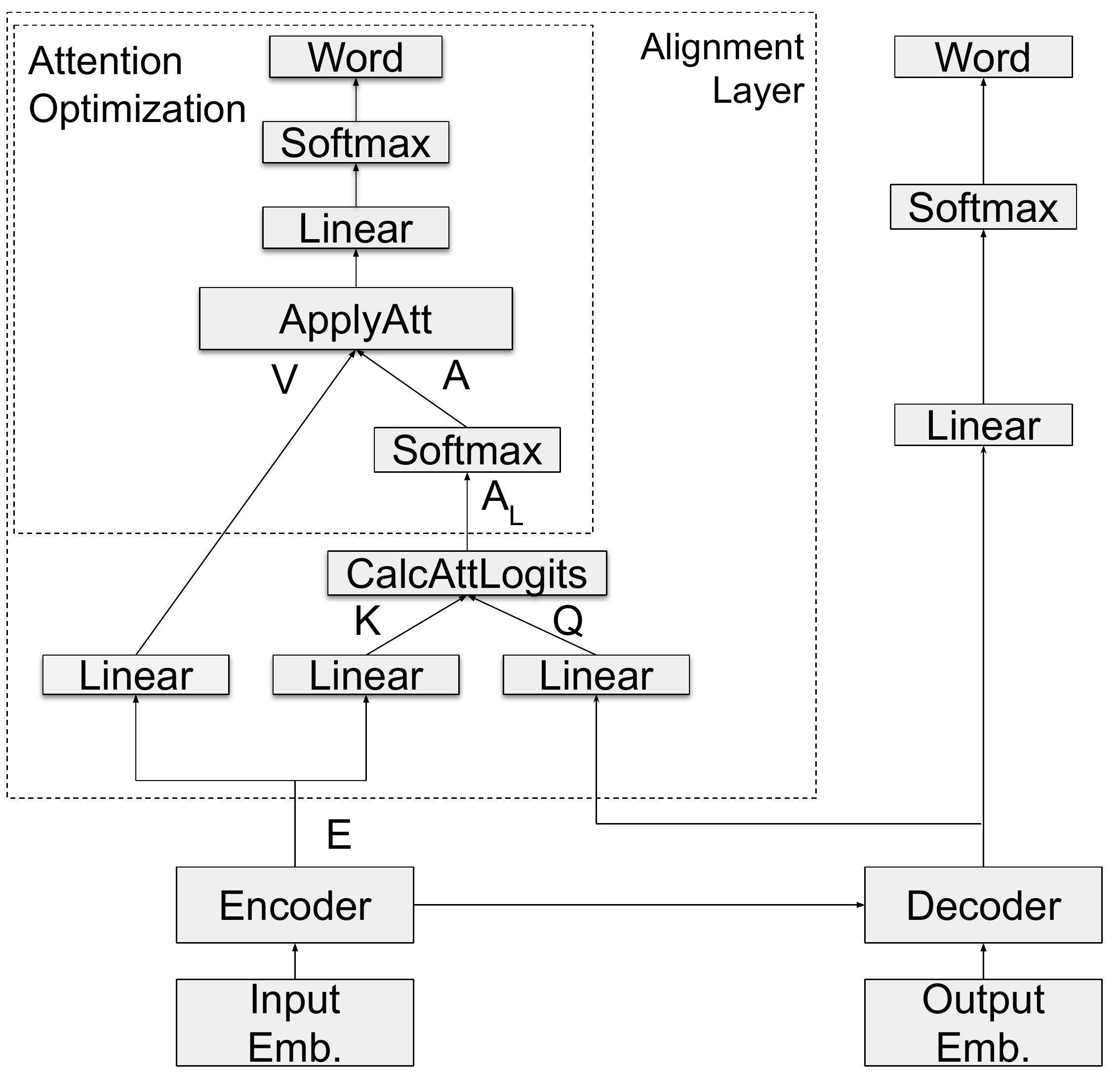}
    \caption{Architecture of the alignment layer. During inference the attention logits $A_L$ of the sub-network \emph{Attention Optimization} are optimized towards predicting the next word correctly.}
    \label{fig:training}
\end{figure}

To improve the alignment quality achieved by interpreting attention activations, \citet{zenkel2019adding} designed an additional \emph{alignment layer} on top of the Transformer architecture. In the alignment layer, the context vector is computed as $\mathsf{applyAtt}(A,V)$, just as in other decoder layers, but this context vector is the only input to predicting the target word via a linear layer and a softmax that gives a probability distribution over the target vocabulary. This design forces attention onto the source positions that are most useful in predicting the target word. Figure \ref{fig:training} depicts its architecture. 

This alignment layer uses the learned representations of the underlying translation model. 
Alignments can be extracted from the activations of this model by running a forward pass to obtain the attention weights $A$ from the alignment layer and subsequently selecting the maximum probability source position for each target position as an alignment link:
$\left\{(\mathsf{argmax}_i\left(A_{i,j} \right),j) :  j \in [1,m] \right\}$.

The alignment layer predicts the next target token $y_i$ based on the source representations $x$ extracted from the encoder of the Transformer and all past target representations $y_{<i}$ extracted from the decoder. Thus the probability is conditioned as $p(y_i | x, y_{<i})$.
The encoder representation used as key and value for the attention component is the sum of the input embeddings and the encoder output. This ensures that lexical and context information are both salient in the input to the attention component.

\subsection{Attention Optimization}
\label{sec:attOpt}
Extracting alignments with attention-based models works well when used in combination with greedy translation inference \citep{li-etal-2019-word}.
However, the alignment task involves predicting an alignment between a sentence and an observed translation, which requires forced decoding. When a token in the target sentence is unexpected given the preceding target prefix, attention activations computed during forced decoding are not reliable because they do not explicitly condition on the target word being aligned. 

\citet{zenkel2019adding} introduce a method called \emph{attention optimization}, which searches for attention activations that maximize the probability of the output sequence by directly optimizing the attention activations $A$ in the alignment layer using gradient descent for the given sentence pair $(x, y)$ to maximize the probability of each observed target token $y_i$ while keeping all other parameters of the neural network $M$ fixed:
\begin{align*}
    \mathsf{argmax}_{A}\ p(y_i | y_{<i}, x, A ; M)
\end{align*}
Attention optimization yields superior alignments when used during forced decoding 
when gradient descent is initialized with the activations from a forward pass through the alignment layer.

\subsection{Full Context Model with Guided Alignment Loss}
The models described so far are based on autoregressive translation models, so they are limited to only attend to the left context of the target sequence.
However, for the word alignment task the current and future target context is also available and should be considered at inference time.
\citet{garg-etal-2019-jointly} train a single model to both predict the target sentence and the alignments using guided alignment training. 
When the model is trained to predict alignments, the full target context can be used to obtain improved alignment quality.

\begin{figure}
    \centering
    \includegraphics[scale=0.36]{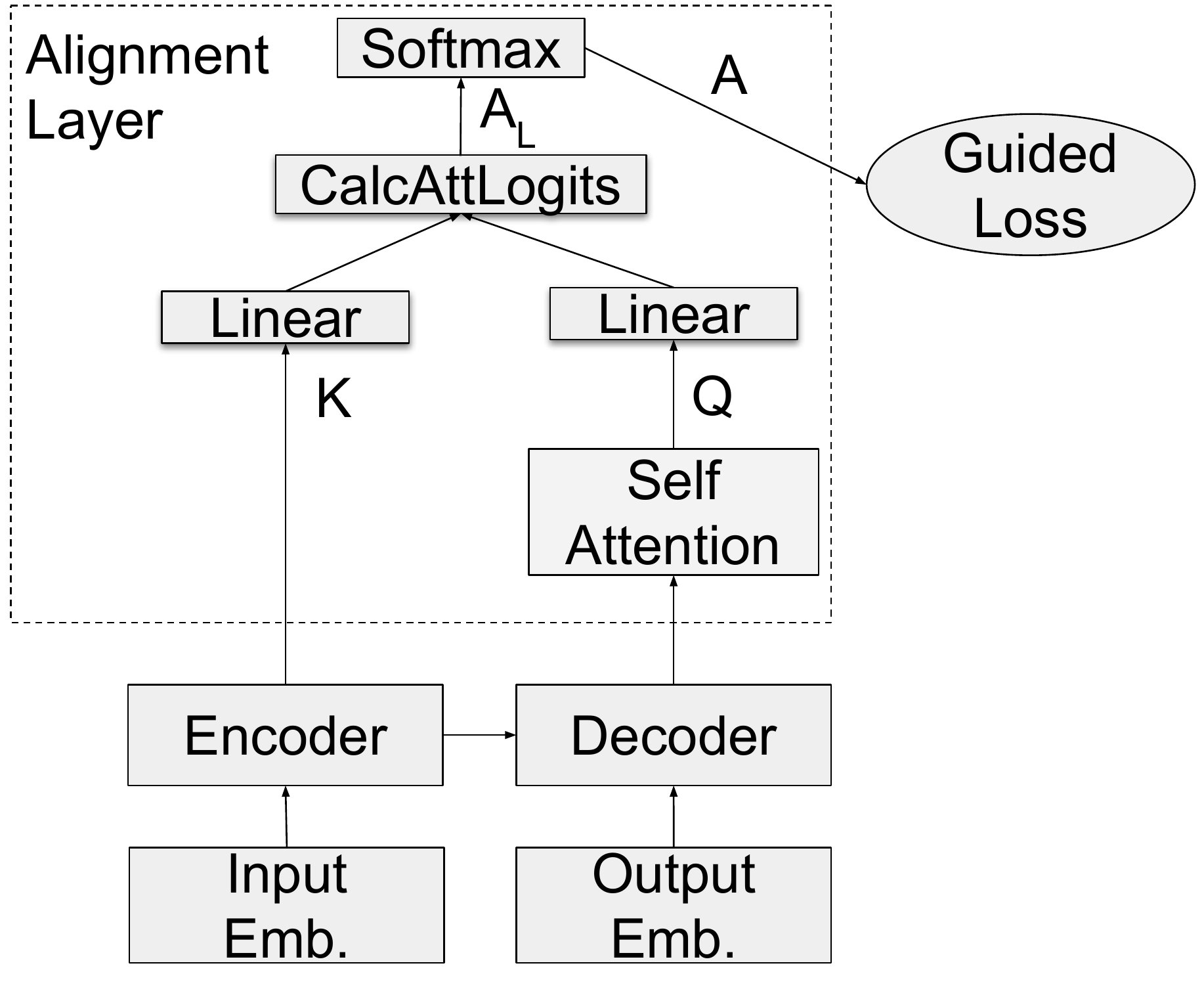}
    \caption{Alignment layer with additional unmasked self attention sublayer to use the full decoder context. 
    }
    \label{fig:guidedLayer}
\end{figure}

The alignment loss requires supervision by a set of alignment links for each sentence pair in the training data. 
These alignments can be generated by the current model or can be provided by an external alignment system or human annotators. 
Assuming one alignment link per target token, we denote the alignment source position for the target token at position $t$ as $a_t$.\footnote{For the purpose of the guided alignment loss we assume target tokens that do not have an alignment link to be aligned to the end-of-sentence (EOS) token of the source sequence.}
The guided alignment loss $L_a$, given attention probabilities $A_{a_t, t}$ for each source position $a_t$ and target position $t$ for a target sequence of length $m$, is defined as:
\begin{align*}
    L_a(A) = -\frac{1}{m} \sum_{i=1}^{m} \log(A_{a_t, t})
\end{align*}
As depicted in Figure \ref{fig:guidedLayer}, we insert an additional self-attention component into the original alignment layer, and leave the encoder and decoder of the Transformer unchanged. 
In contrast to \citet{garg-etal-2019-jointly}, this design does not require updating any translation model parameters; we only optimize the alignment layer parameters with the guided alignment loss.
Adding an alignment layer for guided alignment training has a small parameter overhead as it only adds a single decoder layer, resulting in an increase in parameters of less than 5\%.\footnote{The translation model contains 15 million parameters, while the additional alignment layer has 700 thousand parameters.
}

Unlike the standard decoder-side self-attention layers in the Transformer architecture, the current and future target context are not masked in the alignment layer self-attention component in order to provide the full target sentence as context.
Alignment layer parameters are trained using the guided alignment loss.

\section{Contiguity Loss}
\label{sec:ContiguityLoss}

Contiguous alignment connections are very common in word alignments, especially for pairs of Indo-European languages. That is, if a target word at position $t$ is aligned to a source word at position $s$, the next target word at position $t+1$ is often aligned to $s-1, s$ or $s+1$ \citep{vogel-etal-1996-hmm}. 

Our goal is to design a loss function that encourages alignments with contiguous clusters of links. 

The attention activations form a 2-dimensional matrix $A \in \mathbb{R}^{n \times m}$,
where $n$ is the number of source tokens and $m$ the number of target tokens: each entry represents a probability that specifies how much attention weight the network puts on each source word to predict the next target word. By using a convolution with a static kernel $K$ over these attention scores, we can measure how much attention is focused on each rectangle within the two dimensional attention matrix:
\begin{align*}
    \bar{A} &= \mathsf{conv}(A, K) \\
    L_C &= - \sum_{t=1}^{m} \log(\max_{s \in \{1, ..., n\}} (\bar{A}_{s, t}))
\end{align*}

We use a $2 \times 2$ kernel $K \in \mathbb{R}^{2 \times 2}$ with each element set to 0.5. Therefore, $\bar{A} \in \mathbf{R}^{n \times m}$ will contain the normalized attention mass of each $2 \times 2$ square of the attention matrix $A$.  The resulting values after the convolution will be in the interval $[0.0, 1.0]$. For each target word we select the square with the highest attention mass, encouraging a sparse distribution over source positions in $\bar{A}$ and thus effectively training the model towards strong attention values on neighboring positions.
We mask the contiguity loss such that the end of sentence symbol is not considered during this procedure.
We apply a position-wise dropout of 0.1 on the attention logits before using the softmax function to obtain $A$, which turned out to be important to avoid getting stuck in trivial solutions during training.\footnote{A trivial solution the network converged to when adding the contiguity loss without dropout was to align each target token to the same source token.} 

\begin{figure}
    \centering
    \includegraphics[scale=0.4]{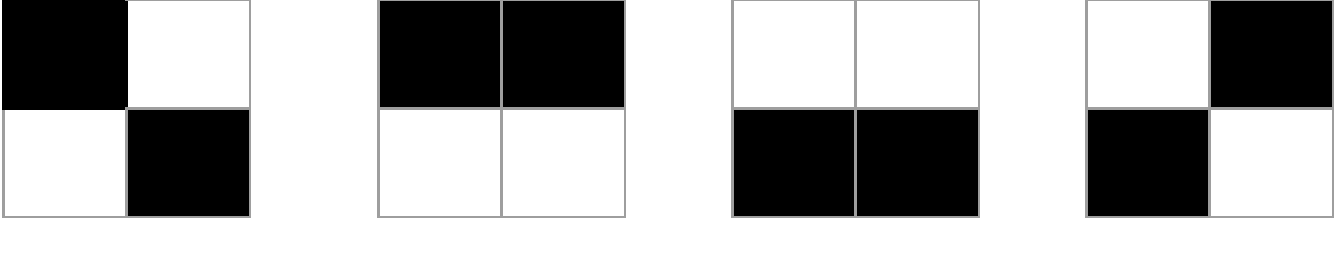}
    \caption{Example of alignment patterns that lead to a minimal contiguity loss.} 
    \label{fig:optAlignments}
\end{figure}

Optimizing the alignment loss especially encourages diagonal and horizontal patterns\footnote{Vertical patterns are not encouraged, as it is not possible to have an attention probability above 0.5 for two source words and the same target word, because we use the softmax function over the source dimension.} as visualized in Figure \ref{fig:optAlignments}. These correspond well to a large portion of patterns appearing in human alignment annotations as shown in Figure \ref{fig:exampleGold}. 

\section{Bidirectional Attention Optimization}
\label{sec:BidirectionAttentionOptimization}
A common way to extract word alignments is to train two models, one for the forward direction (source to target) and one for the backward direction (target to source). For each model, one can extract separate word alignments and symmetrize these using heuristics like grow-diagonal \citep{Och_2000,koehn2005edinburgh}.

\begin{figure}
    \centering
    \includegraphics[scale=0.29]{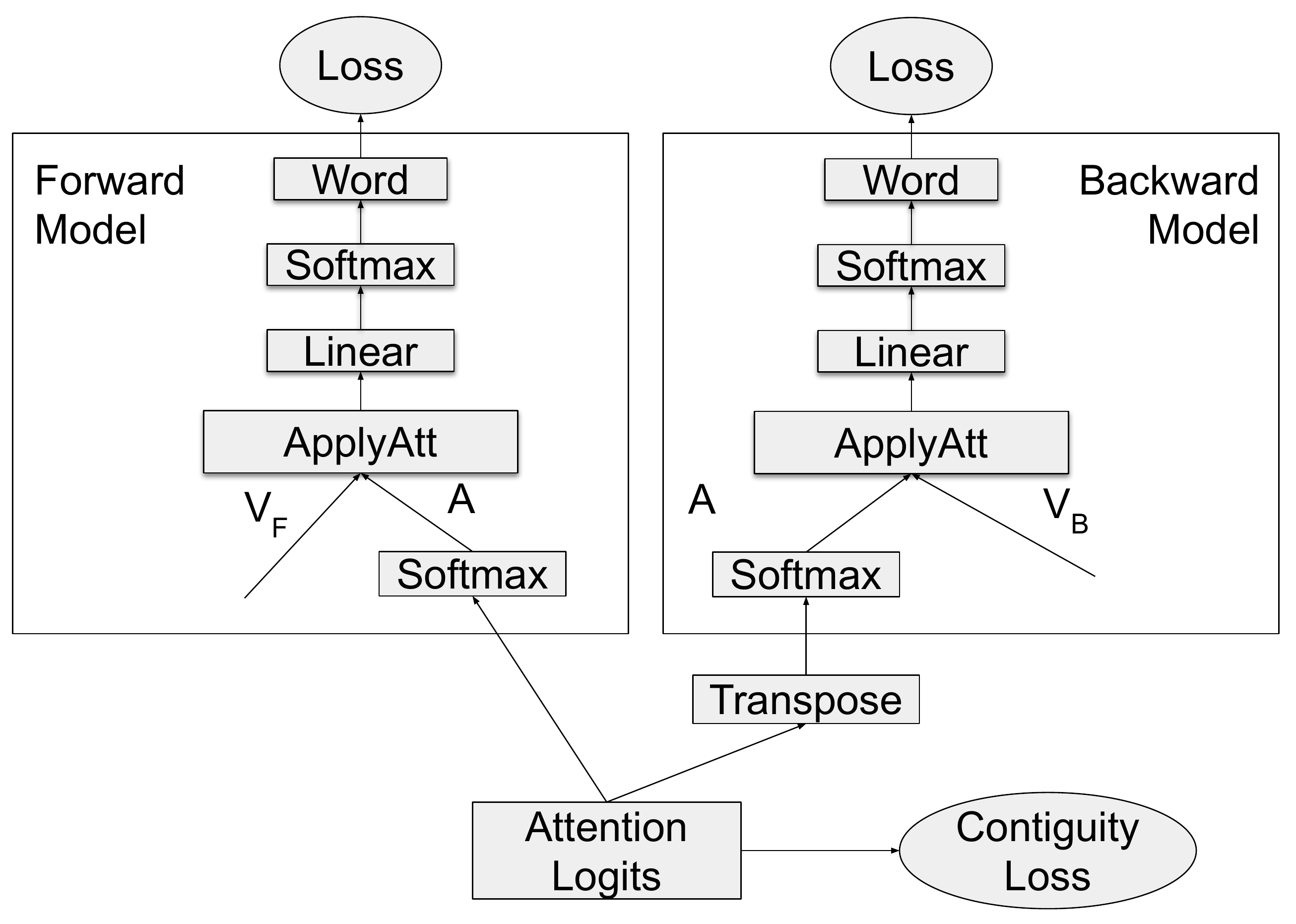}
    \caption{Bidirectional Attention Optimization. We optimize the attention logits towards the correct prediction of the next token when used for both the forward and backward model. The attention values $V_F$ and $V_B$ extracted from the forward and backward model remain static. Additionally, the attention logits are biased towards producing contiguous alignments.}
    \label{fig:bidirOpt}
\end{figure}

However, this approach uses the hard word alignments of both directions as an input, and does not consider any other information of the forward and backward model. For attention-based neural networks it is possible to adapt attention optimization as described in Section \ref{sec:attOpt} to consider two models at the same time. The goal of attention optimization is to find attention activations that lead to the correct prediction of the target sequence for a single neural network. We extend this procedure to optimize the likelihood of the sentence pair jointly under both the forward and the backward model, with the additional bias to favor contiguous alignments. Figure~\ref{fig:bidirOpt} depicts this procedure.

\subsection{Initialization}
Since attention optimization uses gradient descent to find good attention activations, it is important to start with a reasonable initialization. We extract the attention logits (attention before applying the softmax) from the forward $(A_L)_F$ and the backward model $(A_L)_B$ and average these to get a starting point for gradient descent: $(A_L)_{init}~=~\frac{1}{2}((A_L)_F+~(A_L)_B^T)$.

\subsection{Optimization}
Our goal is to find attention logits $A_L$ that lead to the correct prediction for both the forward $M_F$ and the backward model $M_B$, while also representing contiguous alignments. We will use the cross entropy loss $\mathsf{CE}$ for a whole target sequence $y$ of length $m$ to define the loss, given probabilities for each target token $p(y_t|A_t ; M)$ under model parameters $M$ and a given attention activation vector $A_t$:
\begin{align*}
    \mathsf{CE}(p(y|A ; M)) = \sum_{t=1}^{m} -\log (p(y_t|A_t ; M))
\end{align*}

Let $x, y$ be the source and target sequence, so that we can define a loss function for each component with the interpolation parameter $\lambda$ for the contiguity loss $L_C$ as follows:
\begin{align*}
L_F &= \mathsf{CE}(p(y | \mathsf{softmax}(A_L) ; M_F)) \\
L_B &= \mathsf{CE}(p(x | \mathsf{softmax}(A_L^T) ; M_B)) \\
L &= L_F + L_B + \lambda L_C 
\end{align*}
We apply gradient descent to optimize all losses simultaneously, thus approximating a solution of $\mathsf{argmin}_{A_L} L(x,y | A_L, M_F, M_B)$. 

\subsection{Alignment Extraction}
After optimizing the attention logits, we still have to decide which alignment links to extract, i.e. how to convert the soft attentions into hard alignments. For neural models using a single direction a common method is to extract the alignment with the highest attention score for each target token. For our bidirectional method we use the following approach:

We merge the attention probabilities extracted from both directions using element-wise multiplication, where $\otimes$ denotes a Hadamard product:
\begin{align*}
    A_F &= \mathsf{softmax}(A_L) \\
    A_B &= \mathsf{softmax}(A_L^T)^T \\
    A_M &= A_F \otimes A_M
\end{align*}
This favors alignments that effectively predict observed words in both the source and target sentences.

Given the number of source tokens $n$ and target tokens $m$ in the sentence, we select $\min(n, m)$ alignments that have the highest values in the merged attention scores $A_M$. In contrast to selecting one alignment per target token, this allows unaligned tokens, one-to-many, many-to-one and many-to-many alignment patterns.

\section{Experiments}
\label{sec:experiments}

\subsection{Data}
We use the same experimental setup\footnote{\scriptsize\url{https://github.com/lilt/alignment-scripts}} as described by \citet{zenkel2019adding} and used by \citet{garg-etal-2019-jointly}. It contains three language pairs: German$\to$English, Romanian$\to$English and English$\to$French \citep{och2000comparison, Mihalcea_2003}. We learn a joint byte pair encoding (BPE) for the source and the target language with 40k merge operation \citep{sennrich-etal-2016-neural}. To convert from alignments between word pieces to alignments between words, we align a source word to a target word if an alignment link exists between any of its word pieces.

Using BPE units instead of words also improved results for GIZA++ (e.g., 20.9\% vs. 18.9\% for German$\to$English in a single direction). Therefore, we use the exact same input data for GIZA++ and all our neural approaches. For training GIZA++ we use five iterations each for Model 1, the HMM model, Model 3 and Model 4.

\subsection{Training}
Most of the language pairs do not contain an adequately sized development set for word alignment experiments. Therefore, rather than early stopping, we used a fixed number of updates for each training stage across all languages pairs: 90k for training the translation model, 10k for the alignment layer and 10k for guided alignment training (batch-size: 36k words). Training longer did not improve or degrade test-set AER on German$\to$English; the AER only fluctuated by less than 1\% when training the alignment layer for up to 20k updates while evaluating it every 2k updates.

We also trained a base transformer with an alignment layer for German$\to$English, but achieved similar results in terms of AER, so we used the smaller model described in sub-section \ref{subsec:translationModel} for other language pairs. We adopted most hyperparameters from \citet{zenkel2019adding}, see the Supplemental Material for a summary. We tuned the interpolation factor for the contiguity loss on German$\to$English.

\subsection{Contiguity Loss}
Results of ablation experiments for the contiguity loss can be found in Table \ref{tab:contiguityLoss}. Our first experiment uses the contiguity loss during training and we extract the alignments from the forward pass using a single direction without application of attention optimization. We observe an absolute improvement of 6.4\% AER (34.2\% to 27.8\%) after adding the contiguity loss during training.

Afterwards, we use the model trained with contiguity loss and use attention optimization to extract alignments. Adding the contiguity loss during attention optimization further improves the AER scores by 1.2\%. Both during training and attention optimization we used an interpolation coefficient of $\lambda = 1.0$ for the contiguity loss.

\begin{figure*}[t!]
\centering
\begin{subfigure}{.33\textwidth}
  \centering
  \includegraphics[width=1.0\linewidth]{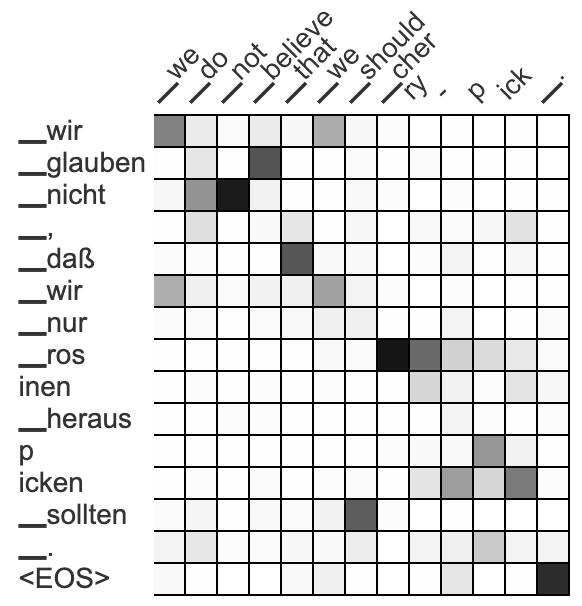}
  \caption{Without Contiguity Loss}
  \label{fig:sub1}
\end{subfigure}%
\begin{subfigure}{.33\textwidth}
  \centering
  \includegraphics[width=1.0\linewidth]{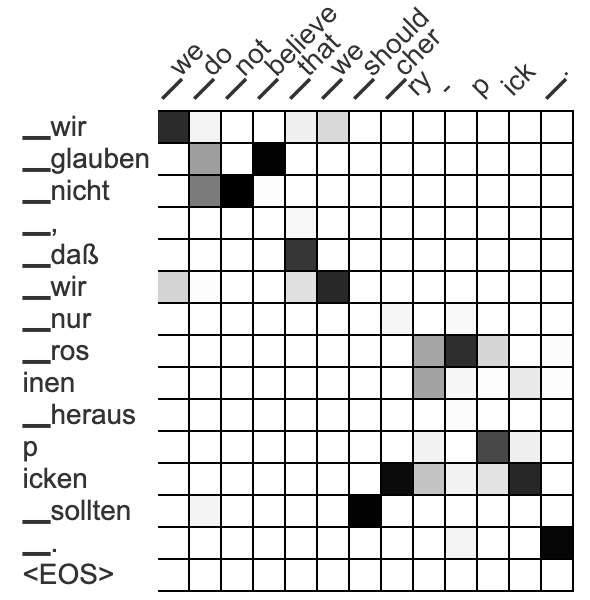}
  \caption{With Contiguity Loss}
  \label{fig:sub2}
\end{subfigure}%
\caption{%
\label{fig:contiguityLoss}%
Attention activations of the alignment layer after attention optimization. Using the contiguity loss during training leads to sparse activations, the correct alignment of the two occurrences of ``we''-``wir'' and to correct alignment of the period.}
\end{figure*}

By visualizing the attention activations in Figure \ref{fig:contiguityLoss} we see that the contiguity loss leads to sparse activations. Additionally, by favoring contiguous alignments it disambiguates correctly the alignment between the words ``we'' and ``wir'', which appear twice in the sentence pair. In the remaining experiments we use the contiguous loss for both training and attention optimization.

While we used a kernel of size 2x2 in our experiments, we also looked at different sizes. Using a 1x1 kernel\footnote{A 1x1 only encourages sparse alignments, and does not encourage contiguous alignments.} during attention optimization leads to an AER of 22.8\%, while a 3x3 kernel achieves the best result with an AER of 21.2\%, compared to 21.5\% of the 2x2 kernel. Larger kernel sizes lead to slightly worse results: 21.4\% for a 4x4 kernel and 21.5\% for a 5x5 kernel. 

\begin{table}[t]
  \centering
  \begin{tabular}{ c  c  c  c }
  Method & No Contiguity & Contiguity \\
  \hline
  Forward & 34.2\% & 27.8\% \\
  Att. Opt & 22.7\% & 21.5\% \\
\end{tabular}
\caption{AER results with and without using the contiguity loss when extracting alignments from the forward pass or when using attention optimization for the language pair German$\to$English.}
\label{tab:contiguityLoss}
\end{table}

\subsection{Bidirectional Attention Optimization}
The most commonly used methods to merge alignments from models trained in opposite directions are variants of grow-diagonal. We extract hard alignments for both German$\to$English and English$\to$German with (monolingual) attention optimization, which leads to an AER of 21.5\% and 25.6\%, respectively. Merging these alignments with grow-diagonal leads to an AER of 19.6\%, while grow-diagonal-final yields an AER of 19.7\%.

We tuned the interpolation factor $\lambda$ for the contiguity loss during bidirectional optimization. A parameter of 1.0 leads to an AER of 18.2\%, 2.0 leads to 18.0\% while 5.0 leads to 17.9\%. Compared to unidirectional attention optimization it makes sense to pick a higher interpolation factor for the contiguity loss, as it is applied with the loss of the forward \textit{and} backward model.

For the remaining experiments we use 5.0 as the interpolation factor. Bidirectional attention optimization improves the resulting alignment error rate compared to the grow-diagonal heuristic by up to 1.8\% for German$\to$English.  These results are summarized in Table \ref{tab:bidirOpt}.

\begin{table}[t]
  \centering
  \begin{tabular}{ l  c }
  & AER \\ \hline
  DeEn & 21.5\% \\
  EnDe & 25.6\% \\ \hline
  Grow-diag &  19.6\% \\
  Grow-diag-final & 19.7\% \\
  Bidir. Att. Opt & \textbf{17.9\%} \\

\end{tabular}
\caption{Comparison of AER scores between bidirectional attention optimization and methods to merge hard alignments.}
\label{tab:bidirOpt}
\end{table}

Variants of grow-diagonal have to rely on the hard alignments generated by the forward and the backward model. They only choose from these alignment links and therefore do not have the ability to generate new alignment links.

\begin{figure*}[t!]
\centering
\begin{subfigure}{.4625\textwidth}
  \centering
  \includegraphics[width=1.0\linewidth]{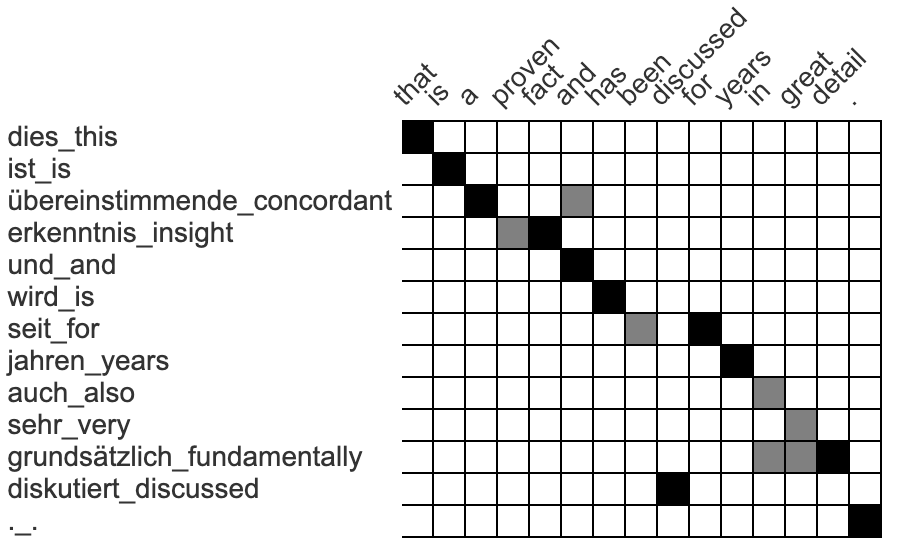}
  \caption{Intersection/Union}
  \label{fig:bidir1}
\end{subfigure}%
\begin{subfigure}{.26\textwidth}
  \centering
  \includegraphics[width=1.0\linewidth]{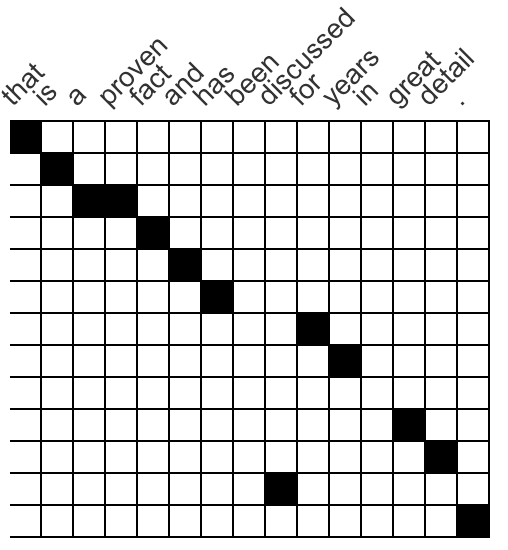}
  \caption{Bidir. Optimization}
  \label{fig:bidir3}
\end{subfigure}%
\begin{subfigure}{.26\textwidth}
  \centering
  \includegraphics[width=1.0\linewidth]{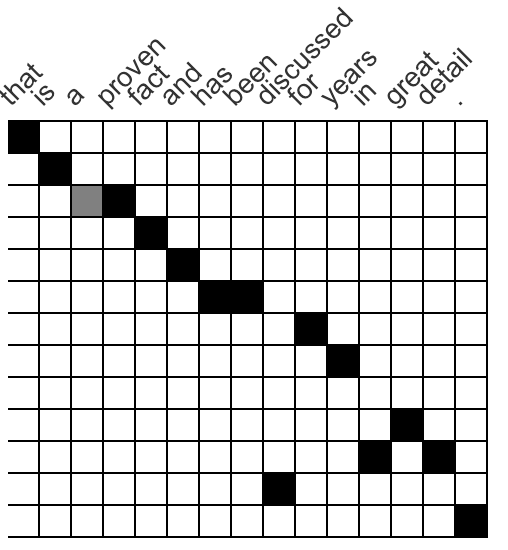}
  \caption{Gold Alignments}
  \label{fig:bidir4}
\end{subfigure}%
\caption{%
\label{fig:bidirOptVis}%
Example of symmetrization with bidirectional attention optimization. We show all alignments extracted from the forward and backward direction with unidirectional attention optimization in Subfigure \ref{fig:bidir1} (alignments that are only present in one direction are grey). Bidirectional attention optimization is able to extract the correct alignment between ``\"ubereinstimmend`` and ``proven'' which did neither appear as an alignment link in the forward nor in the backward direction.}
\end{figure*}

In contrast, bidirectional attention optimization takes the parameters of the underlying models into account and optimizes the underlying attention logits simultaneously for both models to fit the sentence pair. In the example in Figure \ref{fig:bidirOptVis} bidirectional attention optimization is able to correctly predict an alignment link between ``\"ubereinstimmend'' and ``proven'' that did not appear at all in the individual alignments of the forward and backward model.

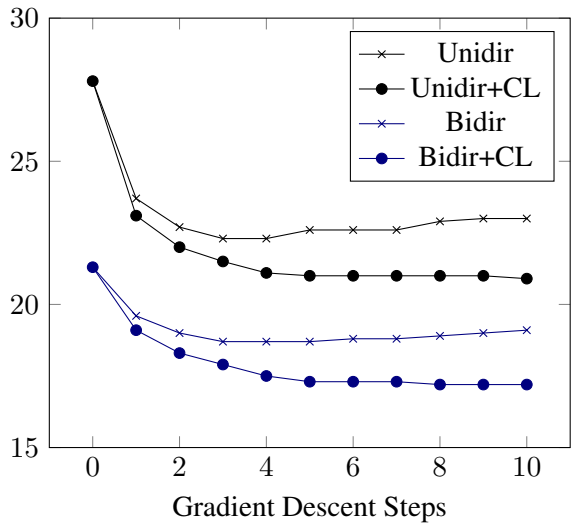
\begin{figure}
\begin{tikzpicture}
	\begin{axis}[
		xlabel=Gradient Descent Steps,
        ymin=15,
        ymax=30,
        legend pos=north east,
]

	\addplot[color=black,mark=x] coordinates {
		(0, 27.8)
		(1, 23.7)
		(2, 22.7)
		(3, 22.3)
		(4, 22.3)
		(5, 22.6)
		(6, 22.6)
		(7, 22.6)
		(8, 22.9)
		(9, 23.0)
		(10, 23.0)
	};
	
	\addplot[color=black,mark=*] coordinates {
		(0, 27.8)
		(1, 23.1)
		(2, 22.0)
		(3, 21.5)
        (4, 21.1)
		(5, 21.0)
		(6, 21.0)
		(7, 21.0)
		(8, 21.0)
		(9, 21.0)
		(10, 20.9)
   };  
	
	\addplot[color=darkblue,mark=x] coordinates {
		(0, 21.3)
		(1, 19.6)
		(2, 19.0)
		(3, 18.7)
		(4, 18.7)
		(5, 18.7)
		(6, 18.8)
		(7, 18.8)
		(8, 18.9)
		(9, 19.0)
		(10, 19.1)
	};
	
	\addplot[color=darkblue,mark=*] coordinates {
		(0, 21.3)
		(1, 19.1)
		(2, 18.3)
		(3, 17.9)
		(4, 17.5)
		(5, 17.3)
		(6, 17.3)
		(7, 17.3)
		(8, 17.2)
		(9, 17.2)
		(10, 17.2)
	};

\legend{Unidir, Unidir+CL, Bidir, Bidir+CL}
\end{axis}
\end{tikzpicture}
\caption{AER with respect to gradient descent steps during attention optimization for German$\to$English. Both unidirectional (Unidir) and bidirectional (Bidir) optimization benefit from the contiguity loss (CL). Without the contiguity loss AER slightly degrades after more than three optimization steps.} 
\label{fig:gdSteps}
\end{figure}

We plot the behavior of attention optimization with a varying number of gradient descent steps in Figure \ref{fig:gdSteps}. For both unidirectional and bidirectional models attention optimization leads to steadily improving results. Without using the additional contiguity loss, the lowest AER appears after three gradient descent steps and slightly increases afterwards. When using the contiguity loss AER results continue to decrease with additional steps. The contiguity loss seems to stabilize optimization and avoids overfitting of the optimized attention activations when tuning them for a single sentence pair.

\subsection{Guided Alignment Training}
We now use the alignment layer with the full decoder context by adding an additional self-attention layer that does not mask out the future target context. We extract alignments from the previous models with bidirectional attention optimization and use those alignments for guided alignment training. 

This works surprisingly well. While the alignments used for training yielded an AER of 17.9\% after bidirectional attention optimization (Table \ref{tab:summarySymmetrization}), the full context model trained with these alignments further improved the AER to 16.0\%  while using a single model for German$\to$English (Table \ref{tab:summarySingleModel}). After guided alignment training is complete, we do not apply attention optimization, since that would require a distribution over target words, which is not available in this model.

\subsection{End-to-End Results}
We now report AER results across all three language pairs. Precision and recall scores are included in the Supplemental Material. We first extract alignments from a unidirectional model, a common use case where translations and alignments need to be extracted simultaneously. Table \ref{tab:summarySingleModel} compares our results to GIZA++ and \citet{zenkel2019adding}.\footnote{\citet{garg-etal-2019-jointly} only report bidirectional results after symmetrization.} We observe that guided alignment training leads to gains across all language pairs. In a single direction our approach consistently outperforms GIZA++ by an absolute AER difference between 1.3\% (EnFr) and 3.9\% (RoEn). 

\begin{table}[t]
  \centering
  \begin{tabular}{ c  c  c  c }
  Method & DeEn & EnFr & RoEn \\
  \hline
  Att. Opt. & 21.5\% & 15.0\% & 29.2\% \\
  +Guided & \textbf{16.0\%} & \textbf{6.6\%} & \textbf{23.4\%} \\
  \hline 
  \citet{zenkel2019adding} & 26.6\% & 23.8\% & 32.3\% \\
  GIZA++ & 18.9\% & 7.9\% & 27.3\% \\
\end{tabular}
\caption{Comparison of unidirectional models with GIZA++.}
\label{tab:summarySingleModel}
\end{table}

\begin{table}[t]
  \centering
  \begin{tabular}{ c  c  c  c }
  Method & DeEn & EnFr & RoEn \\
  \hline
  Bidir. Att. Opt. & 17.9\% & 8.4\% & 24.1\% \\
  +Guided & \textbf{16.3\%} & \textbf{5.0\%} & \textbf{23.4\%} \\
  \hline 
  \citet{zenkel2019adding} & 21.2\% & 10.0\% & 27.6\% \\
  \citet{garg-etal-2019-jointly} & 20.2\% & 7.7\% & 26.0\% \\ 
  GIZA++ & 18.7\% & 5.5\% & 26.5\% \\
\end{tabular}
\caption{Comparison of neural alignment approaches with GIZA++ after using symmetrization of the forward and backward model.}
\label{tab:summarySymmetrization}
\end{table}

Table \ref{tab:summarySymmetrization} compares bidirectional results after symmetrization. We compare to purely neural and purely statistical systems.\footnote{For additional comparisons including neural models bootstrapped with GIZA++ alignments, see the Supplemental Material.} For symmetrizing alignments of the guided model and GIZA++, we use grow-diagonal. Bidirectional attention optimization is already able to outperform GIZA++ and \citet{garg-etal-2019-jointly} on all language pairs except English$\to$French. Using guided alignment training further improves results across all language pairs and leads to a consistent AER improvement compared to GIZA++ and neural results reported by \citet{garg-etal-2019-jointly}.

These results show that it is possible to outperform GIZA++ both in a single direction and after symmetrization without using any alignments generated from statistical alignment systems to bootstrap training.

\section{Conclusion}
\label{sec:Conclusion}

This work presents the first end-to-end neural approach to the word alignment task which consistently outperforms GIZA++ in terms of alignment error rate.
Our approach extends a pre-trained state-of-the-art neural translation model with an additional alignment layer, which is trained in isolation without changing the parameters used for the translation task.
We introduce a novel auxiliary loss function to encourage contiguity in the alignment matrix and a symmetrization algorithm that jointly optimizes the alignment matrix within two models which are trained in opposite directions. 
In a final step the model is re-trained to leverage full target context with a guided alignment loss.
Our results on three language pairs are consistently superior to both GIZA++ and prior work on end-to-end neural alignment.
As the resulting model repurposes a pre-trained translation model without changing its parameters, it can directly benefit from improvements in translation quality, e.g. by adaptation via fine-tuning.


\bibliography{anthology,acl2020}
\bibliographystyle{acl_natbib}

\newpage
\clearpage

\appendix


\section{Supplemental Material}
\label{sec:supplemental}

Table \ref{tab:hypTranslation} and Table \ref{tab:hypAlignmentLayer} summarize the hyperparameters used for the translation model and the additional alignment layer. In Table \ref{tab:allAppendix} we report both AER results and precision and recall for all language pairs.

\begin{table}[htp]
  \centering
  \begin{tabular}{ l  c }
  Hyperparameter & Value \\
  \hline
  Dropout Rate & 0.1 \\
  Embedding Size & 256 \\
  Hidden Units & 512 \\
  Encoder Layers & 6 \\
  Decoder Layers & 3 \\
  Attention Heads Per Layer & 8 \\
\end{tabular}
\caption{Hyperparameters of the translation model.}
\label{tab:hypTranslation}
\end{table}

\begin{table}[htp]
  \centering
  \begin{tabular}{ l  c }
  Hyperparameter & Value \\
  \hline
  Dropout Rate & 0.1 \\
  Embedding Size & 256 \\
  Hidden Units & 256 \\
  Attention Heads & 1 \\
\end{tabular}
\caption{Hyperparameters of the alignment layer.}
\label{tab:hypAlignmentLayer}
\end{table}

\begin{table*}
  \centering
  \begin{tabular}{ l || c  c  c | c  c  c  | c  c  c}
  Method & DeEn & EnDe & Bidir & EnFr & FrEn & Bidir & RoEn & EnRo & Bidir \\
  \hline
  \hline
  \multirow{2}{*}{Att. Opt.} & 21.5\% & 25.6\% & 17.9\% & 15.0\% & 14.3\% & 8.4\% & 29.2\% & 28.8\% & 24.1\% \\
  & 76/81 & 73/76 & 85/79 & 81/92 & 82/93 & 90/95 & 74/68 & 74/69 & 85/69 \\
  \hline
  \multirow{2}{*}{Guided} & 16.0\% & 16.6\% & 16.3\% & 6.6\% & 6.3\% & 5.0\% & 23.4\% & 23.1\% & 23.4\% \\
                          & 88/80  & 89/78  & 93/76  & 92/95  & 93/95  & 96/94 & 88/68 & 90/67  & 93/65 \\
  \hline
  \multirow{2}{*}{GIZA++ (word)} & 20.9\% & 23.1\% & 21.4\% & 8.0\% & 9.8\% & 5.9\% & 28.7\% & 32.2\% & 27.9\% \\
  & 86/72 & 87/69 & 94/67 & 91/93 & 92/88 & 98/90 & 83/63 & 80/59 & 94/59 \\
  \hline
    \multirow{2}{*}{GIZA++ (subword)} & 18.9\% & 20.4\% & 18.7\% & 7.9\% & 8.5\% & 5.5\% & 27.3\% & 29.4\% & 26.5\% \\
  & 89/74 & 88/72 & 95/71 & 92/93 & 93/89 & 98/91 & 85/64 & 83/62 & 93/61 \\
  \hline
  \citet{zenkel2019adding} & 26.6\% & 30.4\% & 21.2\% & 23.8\% & 20.5\% & 10.0\% & 32.3\% & 34.8\% & 27.6\% \\
  \hline
  \citet{garg-etal-2019-jointly} & n/a & n/a & 20.2\% & n/a & n/a & 7.7\% & n/a & n/a & 26.0\% \\
  \hspace{2ex} + GIZA++ & n/a & n/a & 16.0\% & n/a & n/a & 4.6\% & n/a & n/a & 23.1\% \\
  \end{tabular}
\caption{AER and---when available---precision/recall scores in percentage in the following row. The \emph{Bidir} column reports results 
for the DeEn, EnFr and RoEn translation direction, respectively, and uses grow-diagonal for all columns except when attention optimization is used. For attention optimization we merge alignments with bidirectional attention optimization.}
\label{tab:allAppendix}
\end{table*}

\end{document}